\documentclass[journal,twoside,web]{ieeecolor}
\usepackage{tmi}
\usepackage{cite}
\usepackage{amsmath,amssymb,amsfonts}
\usepackage{graphicx}
\usepackage{textcomp}
\usepackage{algpseudocode}
\usepackage{multirow}
\usepackage{epstopdf}
\usepackage{amsmath}
\usepackage{lipsum} % For dummy text
\usepackage{float} % For positioning tables
\usepackage{subcaption}
\usepackage[utf8]{inputenc}
\usepackage{mdframed}
\usepackage[ruled,boxed]{algorithm2e}
\usepackage{xcolor}
\usepackage{color}
\definecolor{bottomcolor}{RGB}{247, 248, 236}

\def\BibTeX{{\rm B\kern-.05em{\sc i\kern-.025em b}\kern-.08em
    T\kern-.1667em\lower.7ex\hbox{E}\kern-.125emX}}
\begin{document}
\title{CATD: Unified Representation Learning\\
 for EEG-to-fMRI Cross-Modal Generation}
\author{Weiheng Yao, Zhihan Lyu, Mufti Mahmud, Ning Zhong, Baiying Lei, Shuqiang Wang
\thanks{Weiheng Yao is with Southern University of Science and Technology, Shenzhen, 518055, China.}
\thanks{Weiheng Yao, and Shuqiang Wang are with Shenzhen Institutes of Advanced Technology, Chinese Academy of Sciences, Shenzhen 518055, China}
\thanks{Zhihan Lyu is with Department of Game Design, Faculty of Arts, Uppsala University, Sweden}
\thanks{Mufti Mahmud is with Department of Information and Computer Science , SDAIA-KFUPM Joint Research Center for AI, Interdisciplinary Research Center for Biosystems and Machines, King Fahd University of Petroleum and Minerals, Dhahran, Saudi Arabia}
\thanks{Ning Zhong is with the Faculty of Engineering, Maebashi Institute of Technology, Gunma, Japan and he  is also with Chongqing University of Posts and Telecommunications,Chongqing, China.}
\thanks{Baiying Lei is with School of Biomedical Engineering, Health Science Center, Shenzhen University, Guangdong Key Laboratory for Biomedical Measurements and Ultrasound Imaging, Shenzhen, 518000, Guangdong, China}
\thanks{\textcopyright{}2025 IEEE. Personal use of this material is permitted. Permission from IEEE must be obtained for all other uses, in any current or future media, including reprinting/republishing this material for advertising or promotional purposes, creating new collective works, for resale or redistribution to servers or lists, or reuse of any copyrighted component of this work in other works.}
}

\maketitle

\begin{abstract}
Multi-modal neuroimaging analysis is crucial for a comprehensive understanding of brain function and pathology, as it allows for the integration of different imaging techniques, thus overcoming the limitations of individual modalities. However, the high costs and limited availability of certain modalities pose significant challenges. To address these issues, this paper proposes the Condition-Aligned Temporal Diffusion (CATD) framework for end-to-end cross-modal synthesis of neuroimaging, enabling the generation of functional magnetic resonance imaging (fMRI)-detected Blood Oxygen Level Dependent (BOLD) signals from more accessible Electroencephalography (EEG) signals. By constructing Conditionally Aligned Block (CAB), heterogeneous neuroimages are aligned into a latent space, achieving a unified representation that provides the foundation for cross-modal transformation in neuroimaging. The combination with the constructed Dynamic Time-Frequency Segmentation (DTFS) module also enables the use of EEG signals to improve the temporal resolution of BOLD signals, thus augmenting the capture of the dynamic details of the brain. Experimental validation demonstrates that the framework improves the accuracy of brain activity state prediction by 9.13\% (reaching 69.8\%), enhances the diagnostic accuracy of brain disorders by 4.10\% (reaching 99.55\%), effectively identifies abnormal brain regions, enhancing the temporal resolution of BOLD signals. The proposed framework establishes a new paradigm for cross-modal synthesis of neuroimaging by unifying heterogeneous neuroimaging data into a latent representation space, showing promise in medical applications such as improving Parkinson's disease prediction and identifying abnormal brain regions.
\end{abstract}

\begin{IEEEkeywords}
 Cross-Modal Generation, Representation Learning, Diffusion Model, Functional Neuroimaging, Temporal Super-Resolution
\end{IEEEkeywords}

\section{Introduction}
\label{sec:introduction}

\IEEEPARstart{T}{he} BOLD signal, measured by fMRI, provides a detailed and precise mapping of brain activity\cite{1_kronemer2022human}. The sensitivity of this signal to changes in oxygenation and deoxygenation levels in the blood provides a dynamic image of brain function. This helps to understand and track various brain diseases and is considered the gold standard of modern functional neuroimaging\cite{2_anwar2016effective}. Notably, BOLD fMRI has provided important insights into the brain dynamics of disorders such as ischemic stroke\cite{3_hou2023deep}, schizophrenia\cite{4_braun2021brain}, Alzheimer’s disease\cite{5_gonneaud2021accelerated}, focal epilepsy\cite{6_zijlmans2019changing}, and depression\cite{7_talishinsky2022regional}. This highlights its indispensable role in diagnosing and monitoring these diseases.

\begin{figure}
	\centerline{\includegraphics[width=\linewidth]{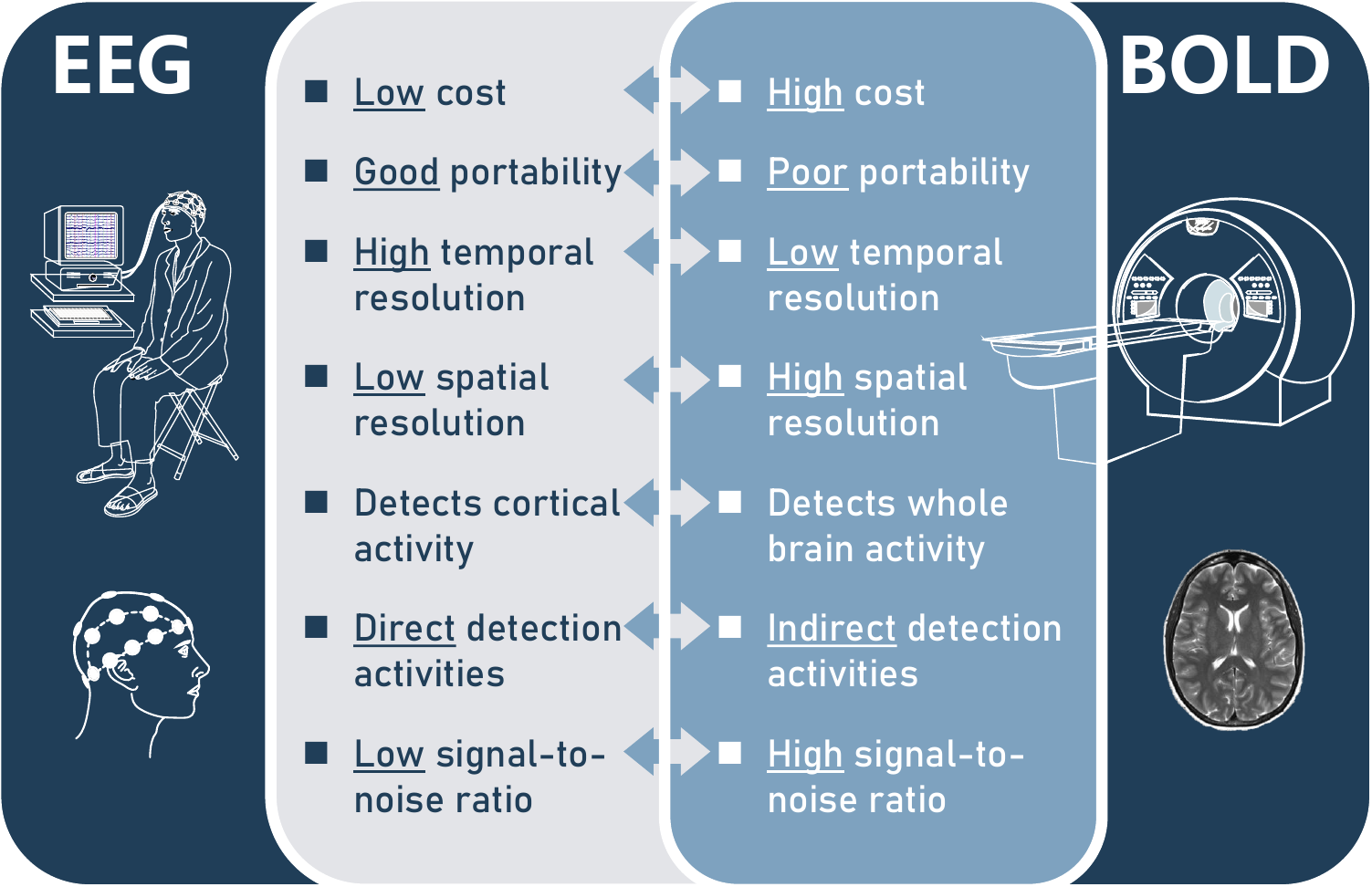}}
	\caption{Comparison of Advantages and Disadvantages of BOLD fMRI and EEG}
	\label{fig1}
\end{figure}

Although BOLD fMRI is highly regarded for its detailed imaging capabilities, acquiring such scans is not only costly\cite{8_demene2017functional} and time-consuming\cite{9_st2023brain} but also subject to limitations in various clinical applications\cite{10_cruse2011bedside, 11_keynan2019electrical}.
Specifically, its poor temporal resolution limits the ability to accurately capture rapid neural dynamics that occur within milliseconds. The hemodynamic response delay inherent to fMRI makes it less suitable for real-time applications, such as brain-machine interfaces or neurofeedback systems\cite{12_casagrande2023comprehensive}. At the same time, the high costs and limited portability of fMRI make it inaccessible for many patients, particularly in low-resource settings. In contrast, EEG has several advantages in reflecting brain activity. Because EEG directly measures brain electrical activity, it provides real-time insight with high temporal resolution, which is critical for capturing rapid dynamic changes in neuronal circuits\cite{13_luo2024electroencephalography}. However, EEG's low spatial resolution significantly hinders its ability to localize brain activity accurately, limiting its use in tasks requiring precise spatial mapping\cite{14_fang2022noninvasive}. The low signal-to-noise ratio of EEG data and its susceptibility to artifacts (e.g., muscle movement, eye blinks) also pose challenges for reliable interpretation, especially in complex cognitive or clinical scenarios. Fig.\ref{fig1} outlines the respective strengths and limitations of fMRI and EEG.

In recent years, with the advancements in information technology\cite{add_1, add_3} and artificial intelligence techniques\cite{hotspot_lyu2024multi, add_2, add_4, add_5}, cross-modal neuroimage synthesis has gradually become an active area of research, especially with the wide application of generative models in different fields. Generative AI models, such as transformer\cite{trasformer_10198494}, diffusion models\cite{DDPM_ho2020denoising, diffusion2_10167641, diffusion3_10288274} and generative adversarial networks (GANs)\cite{GAN1_hu2021bidirectional, GAN2_hu20233, GAN3_10087263}, demonstrate great potential in cross-modal data synthesis\cite{_14_deng2019irc,_15_takagi2023high}, and their applications in brain imaging\cite{genAIinbrain_2} and brain function\cite{genAIinbrain_1, genAIinbrain_3} are also gradually increasing. For example, Yan et al. used a deep convolutional generative adversarial network (DCGAN) to reconstruct missing BOLD signals for individual participants\cite{lhs_yan2020reconstructing}. Calhas et al.\cite{calhas} conducted early research on generative AI for EEG-to-fMRI synthesis, proposing an Autoencoder(AE)-based model for this purpose. In our experiments, we compared our proposed method to their AE-based model and demonstrated significant performance improvements in cross-modal synthesis. The success of generative AI is based on the existence of some of the same underlying information and potentially relevant features in different modalities. Although EEG and fMRI are acquired in different ways, they both reflect brain activity and EEG has the advantage of low cost and fewer limitations on its use. It has been found that there is a strong correlation between microstate transitions in EEG signals and BOLD signals\cite{_16_al2022canonical} and that BOLD functional connectivity correlates with functional connectivity of EEG activity\cite{_SEEG&BOLD_huang2023intracranial}. By simultaneously recording EEG and fMRI signals, researchers can observe the relationship between EEG activity and metabolic activity in the brain. When electrical brain activity increases, blood oxygen levels in the brain also increase, suggesting a one-to-one correspondence between changes in electrical brain activity and the state of metabolic activity in the brain\cite{_17_1_kwon2023brain,_17_2_heeger2002does}. This correlation provides a theoretical basis for cross-modal generation and temporal resolution enhancement of fMRI using generative AI and EEG.

EEG and BOLD signals are highly heterogeneous time-series data, requiring powerful generative AI models for cross-modal synthesis. Currently, diffusion transformer model\cite{DiT_peebles2022scalable} has demonstrated excellent performance in several areas such as image generation. The emergence of Sora\cite{sora_liu2024sora} further demonstrates the potential application of diffusion models for temporal data synthesis. Drawing inspiration from these advancements, the CATD framework for the unified representation of EEG and BOLD signals is proposed. This framework is the first to achieve cross-modal synthesis of high-dimensional, heterogeneous brain functional data using a diffusion model. The novelties and contributions of this paper can be summarized in the following points:

\begin{itemize}
    \item [(i)] A new paradigm based on generative AI for unified representation of neuroimaging is proposed. As far as we know, it is the first time a diffusion-driven end-to-end framework is developed for EEG-to-fMRI synthesis. Utilizing low-cost, accessible EEG signals, the proposed CATD framework is capable of synthesizing high-cost, difficult-to-access BOLD signals. The proposed framework enables high-quality, stable cross-modal generation from EEG to BOLD signals, bridging the gap between different neuroimaging modalities.
    \item [(ii)] The CAB module is designed to align high-temporal, low-spatial resolution EEG signals with low-temporal, high-spatial resolution BOLD signals within a latent space, facilitating a unified representation across modalities. The combination with DTFS module also leverages EEG's superior temporal resolution to improve the temporal super-resolution of BOLD signals, capturing detailed brain dynamics that outperform traditional methods.
\end{itemize}

The remainder of this paper is organized as follows. Section \ref{section2} introduces the proposed CATD framework in detail, including its structural design and methodology. Section \ref{section3} describes the experimental setup, datasets used, implementation details, and the experimental results. Section \ref{section4} provides a comprehensive analysis and discussion of the experimental results as well as the reasons for their occurrence, the applicability and potential of the framework. Finally, Section \ref{section5} concludes the paper by summarizing the key contributions and highlighting future research directions.

\section{Method}\label{section2}

\subsection{Overview}

\begin{figure*}
\centerline{\includegraphics[width=0.9\linewidth]{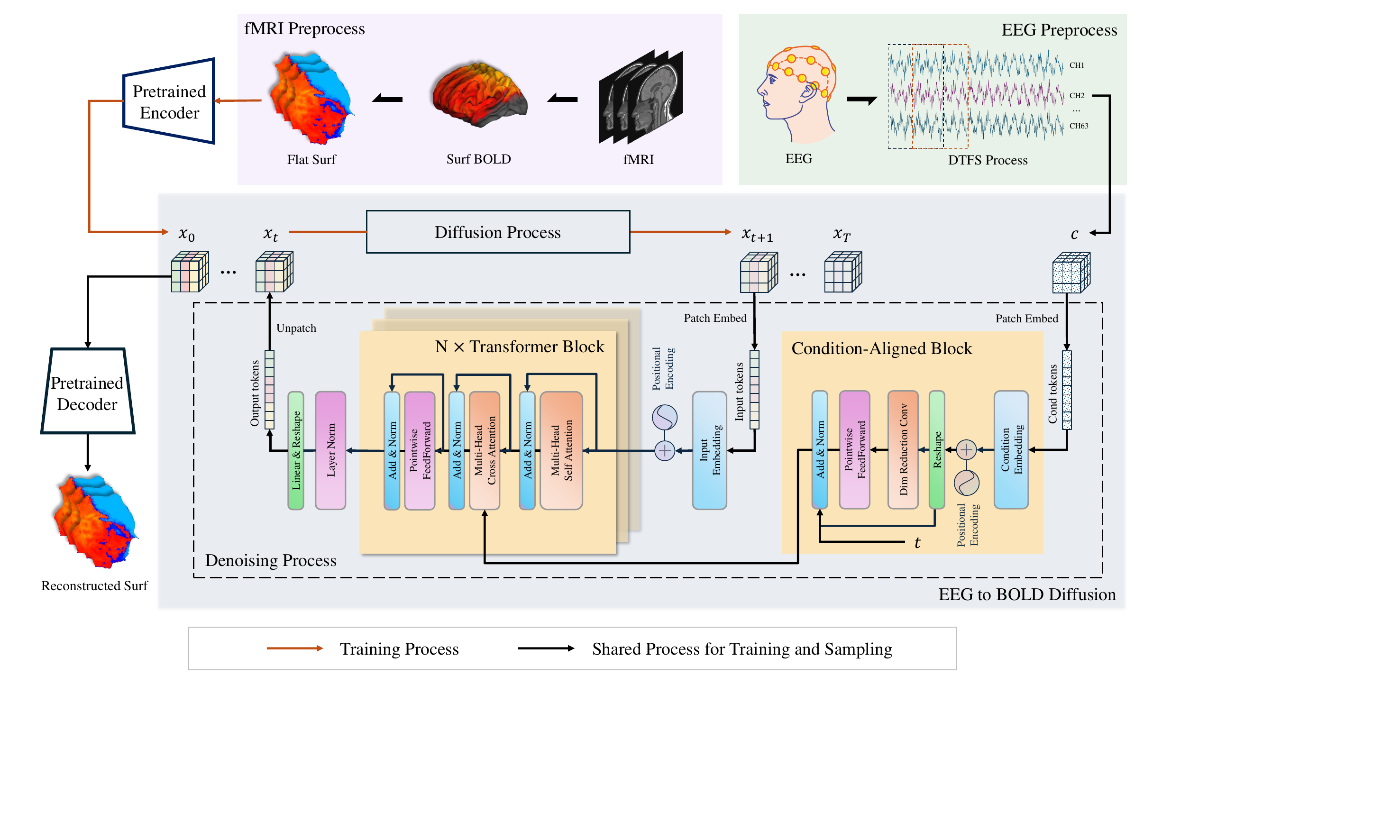}}
\caption{The overall framework of the proposed CATD. The upper part of the figure shows how two different dimensions of data are processed to achieve the initial alignment. The lower half shows the generation pipeline of the BOLD signal under the control of the EEG condition based on the DiT structure.}
\label{fig2}
\end{figure*}

BOLD signals are valuable in reflecting brain activity and are essential for analysing, diagnosing and treating brain disorders. However, BOLD signal acquisition is not possible in some patients due to medical conditions or other limitations. In response, the paper proposes the CATD framework, an innovative EEG-to-BOLD signal conversion model based on Scalable Diffusion Models with Transformers (DiT)\cite{DiT_peebles2022scalable}. This approach addresses the limitations of cross-modal generation and temporal super-resolution of whole-brain BOLD signals, which are not possible with existing techniques. The CATD framework addresses the challenges posed by high dimensionality and asymmetry between EEG and BOLD data through a novel heterogeneous alignment method that facilitates dimensional matching and enhances signal compatibility. At the same time, it employs a cross-attention mechanism to efficiently generate cross-modal EEG-modulated BOLD signals. The proposed DTFS achieves the control of the EEG sampling rate by sliding the sampler, which in turn achieves the enhancement of the temporal resolution of the output BOLD signal. Fig.\ref{fig2} depicts the complete structure and functionality of the CATD framework.

\subsection{Data Alignment Method}

EEG and BOLD signals are high-dimensional signals, one with high temporal resolution and the other with high spatial resolution, highly heterogeneous in scale, and both need to be processed to the same scale using the following methods to perform integration operations. For preprocessing BOLD signals, we followed the method by Yan et al. \cite{lhs_yan2020reconstructing}, utilizing freesurfer\cite{FreeSurfer} software to register brain BOLD activity to the cerebral cortex. Individual structural images were processed to generate cortical surface meshes, and structural and functional images were aligned using boundary-based registration with FsFast. The fMRI data were aligned to the fsaverage template and subsequently downsampled to the fsaverage4 template, creating a functional map with 2562 vertices per hemisphere using the mri\_surf2surf function in FreeSurfer. Additionally, a 6-mm full-width at half-maximum (FWHM) Gaussian smoothing kernel was applied to the fMRI data in surface space. These processed signals are fed into the encoder of a pretrained Variational Autoencoder (VAE)\cite{VAE_kingma2013autoencoding}, specifically sd-vae-ft-mse from the Diffusers library \cite{diffusers}, for dimensionality reduction. The reduced data are then segmented into patches and transformed into input tokens via an embedding layer, producing the initial state $x_0$ for diffusion model training.

EEG signals, corresponding to the first 6 seconds of each BOLD functional map, are selected due to the 6-second delay between neuronal activity and blood oxygenation response \cite{6s1_liao2002estimating, 6s2_burke2013pursuit}. Feature extraction and dimensionality reduction are performed using a dynamic time-frequency analysis method based on the short-time Fourier transform. The reduced EEG data are then divided into segments and converted into tokens that match the BOLD frames in dimension and number through the EEG embedding layer. These tokens serve as the conditional signals $c$ for the diffusion model, enabling alignment of the high-dimensional, heterogeneous time-series EEG data with BOLD signals.

For the BOLD signal super-resolution task, the sampling rate of the EEG signal is controlled using the DTFS module. During training, overlapping EEG signal samples are employed. Specifically, the EEG signal is sampled at one-third of its original interval while maintaining the same overall sampling duration. This fine-grained temporal segmentation approach allows the generated BOLD signal to achieve three times the temporal resolution of the original signal. As a result, this method generates more refined BOLD signal sequences, significantly enhancing temporal resolution and providing a more detailed representation of brain activity.

\subsection{EEG to BOLD Diffusion}

\begin{algorithm}
\caption{EEG to BOLD Signal Generation using Diffusion Model: Training Phase}
\begin{mdframed}[backgroundcolor=bottomcolor,rightline=false,leftline=false,topline=false,bottomline=false,innerleftmargin=5pt,innerrightmargin=5pt,userdefinedwidth=0.92\linewidth,innerbottommargin=5pt,innertopmargin=5pt]

\textbf{Define:}

$F_{BOLD}(x)$: Function to preprocess BOLD data

$F_{EEG}(c, f_s)$: DTFS process, with $f_s$ as sample rate

Encoder$(x)$: Encoder for fMRI data

Decoder$(x)$: Decoder for fMRI data

CAB$(c)$: Condition-Aligned Block function

$\alpha_t, \beta_t$: Noise schedule parameters

$\epsilon_\theta$: Neural network for predicting noise

$\mathcal{L}(\theta)$: Loss function

\BlankLine

\textbf{Input:}

$x$ \tcp*[r]{Raw fMRI data}

$c$ \tcp*[r]{Raw EEG data}

$f_s$ \tcp*[r]{EEG sample rate}

$N$ \tcp*[r]{Number of diffusion steps}

% $\alpha_t, \beta_t$ \tcp*[r]{Noise schedule parameters}

\BlankLine

\textbf{Get BOLD Data:}

$x \gets$  Load raw fMRI data

$x \gets F_{BOLD}\left(x\right)$

$x \gets Encoder\left(x\right)$

\BlankLine

\textbf{Initialize Diffusion Model:}

$x_0 \gets x$ \tcp*[r]{Initialize with preprocessed BOLD data}

\BlankLine

\textbf{Forward Diffusion Process:}

\For{$t \gets 1$ \KwTo $N$}
{
    Sample $\epsilon_t \sim \mathcal{N}(0, I)$ \;
    $x_t \gets \sqrt{\alpha_t} x_{t-1} + \sqrt{\beta_t} \epsilon_t$
}

\BlankLine

\textbf{Reverse Denoising Process:}

\For{$t \gets N$ \KwTo $1$}
{
    $\hat{\epsilon_t} \gets \epsilon_\theta(x_t, CAB(c))$ \;
    $x_{t-1} \gets \frac{1}{\sqrt{\alpha_t}} \left( x_t - \frac{\beta_t}{\sqrt{1 - \bar{\alpha}_t}} \hat{\epsilon_t} \right)$
}

\BlankLine

\textbf{Loss Calculation and Backpropagation:}

$\mathcal{L}(\theta) = {\left \| \epsilon_t - \epsilon_\theta(x_t, CAB(c)) \right \|}^2_2$

\BlankLine

\textbf{Update model parameters using backpropagation}

\BlankLine

\end{mdframed}
\end{algorithm}

\begin{algorithm}
\caption{EEG to BOLD Signal Generation using Diffusion Model: Inference Phase}
\begin{mdframed}[backgroundcolor=bottomcolor,rightline=false,leftline=false,topline=false,bottomline=false,innerleftmargin=5pt,innerrightmargin=5pt,userdefinedwidth=0.92\linewidth,innerbottommargin=5pt,innertopmargin=5pt]

\BlankLine

\textbf{Initialize Diffusion Model for Inference:}

$x_T \gets  Initialize with noise$

$N \gets Set number of diffusion steps$

\BlankLine

\textbf{Reverse Denoising Process:}

\For{$t \gets N$ \KwTo $1$}{
    $\hat{\epsilon_t} \gets \epsilon_\theta(x_t, CAB(c))$ \;
    $x_{t-1} \gets \frac{1}{\sqrt{\alpha_t}} \left( x_t - \frac{\beta_t}{\sqrt{1 - \bar{\alpha}_t}} \hat{\epsilon_t} \right)$
}

\BlankLine

\textbf{Decode Generated Signal:}

$x \gets Decoder(x_0)$ \tcp*[r]{Decode generated BOLD signal}

\BlankLine

\textbf{Output:}

Reconstruct final BOLD signal from $x$ \;

\BlankLine

\end{mdframed}
\end{algorithm}

\subsubsection{Basic ideas}
As one of the highest performing generative AI models available, diffusion models are known for their stable training process and superior quality of generated output. These models work by gradually transforming the data distribution into a Gaussian distribution through a forward process, and then learning the reverse transformation to generate new data samples. In the forward phase, the diffusion model starts with the original data $x_0$ and gradually increases the noise over multiple time steps, eventually reaching an almost completely random state $x_T$, a process that can be described by the following Gaussian process:
\begin{equation}
    q\left ( x_t|x_0 \right ) = \mathcal{N} \left ( x_t;\sqrt{\bar{\alpha }_t } x_0,\left ( 1-\bar{\alpha }_t \right ) I \right ) ,
    \label{eq1}
\end{equation}
where $\bar{\alpha }_t$ is a hyperparameter that decreases over time. This process is characterised by a Markov chain and each step is typically governed by a variance-preserving transformation.

The inverse process seeks to reconstruct the original data from the noisy state. This is done by training a neural network to predict the noise added at each step of the forward process, and then iteratively removing this noise to recover the original data $x_0$. This inverse process can be described by the following equation:
\begin{equation}
    p_\theta \left ( x_{t-1}|x_t,c \right ) = \mathcal{N} \left ( x_{t-1};\mu _\theta \left ( x_t,c \right ) ,\Sigma _\theta \left ( x_t,c \right )  \right ),
    \label{eq2}
\end{equation}
where $\mu _\theta \left ( x_t,c \right )$ and $\Sigma _\theta \left ( x_t,c \right ) $ are predicted by a neural network with parameter $\theta$. We adopt a transformer architecture for this network because of its flexibility and excellent ability to capture long-range dependencies, which makes it particularly suitable for processing time-series signals such as EEG. The inherent attention mechanism of the transformer allows for the integration of conditional information denoted by $c$ through a cross-attention mechanism. This integration guides the generation process, thereby enhancing the model’s applicability to cross-modal generation tasks involving high-dimensional, heterogeneous data.

\subsubsection{Architectures}
As shown in Fig.\ref{fig2}, our proposed EEG to BOLD diffusion model consists of a prediction network that consists of CAB, several layers of Transformer Blocks connected in series, an input embedding layer, and an output section. Each Transformer Block primarily contains one layer of a multi-head self-attention mechanism and one layer of multi-head cross-attention mechanism. The cross-attention layer is responsible for incorporating the conditional information into the network. The CAB integrates EEG markers as condition $c$ as well as the diffusion step $t, t\in \left \{ 0,1,\dots,T \right \}$, and delivers this integrated conditional information to the cross-attention network of each Transformer Module as a way to achieve an effective input of conditional information.

\subsubsection{Loss Function}
According to Eq.\ref{eq1} and Eq.\ref{eq2}, the inverse process is trained using a log-likelihood variational lower bound on $x_0$, which can be simplified as
\begin{equation}
\begin{split}
    \mathcal{L} \left ( \theta  \right ) = & -p\left ( x_0|x_1 \right ) + \\ &{\textstyle \sum_{t}^{} \mathcal{D} _{KL}\left ( q^*\left ( x_{t-1}|x_t, x_0 \right ) \parallel p_\theta \left ( x_{t-1}|x_t,c \right )  \right ) },
\end{split}
    \label{eq3}
\end{equation}
where $q^*$ denotes is the true conditional distribution of $x_{t-1}$ given $x_t$ and $x_0$, and $\mathcal{D} _{KL}$ denotes the KL dispersion of the two distributions.

In discussing the training process of the diffusion model, it is crucial to consider the consistency of the entire generation process, which requires the model to complete the computation of all time steps $T$ before each parameter update. While this approach ensures the comprehensiveness of the learning process, it has a significant negative impact on the convergence speed, stability and optimisation efficiency of the model. At the same time, it also places a high demand on computational resources. In view of this, we decided to simplify Eq.\ref{eq3}. Instead of minimizing the KL divergence to ensure distributional similarity across all time steps, we instead turned to minimising the prediction error in each time step. This approach not only simplifies the computational process but also helps to improve the training efficiency and stability of the model. The expression for the simplified loss function is as follows:
\begin{equation}
    \mathcal{L} \left ( \theta \right ) ={\left \| \epsilon _t - \epsilon _\theta \left ( x_t,CAB(c) \right )  \right \|}^2_2  ,
    \label{eq4}
\end{equation}
where $\epsilon _\theta \left ( x_t,CAB(c) \right )$ denotes the noise predicted by the network and $\epsilon _t$ denotes the ground truth sampled Gaussian noise.

\section{Experiment}\label{section3}

\subsection{Experiment Settings}

\begin{figure*}
	\centering
	\includegraphics[width=0.7\linewidth]{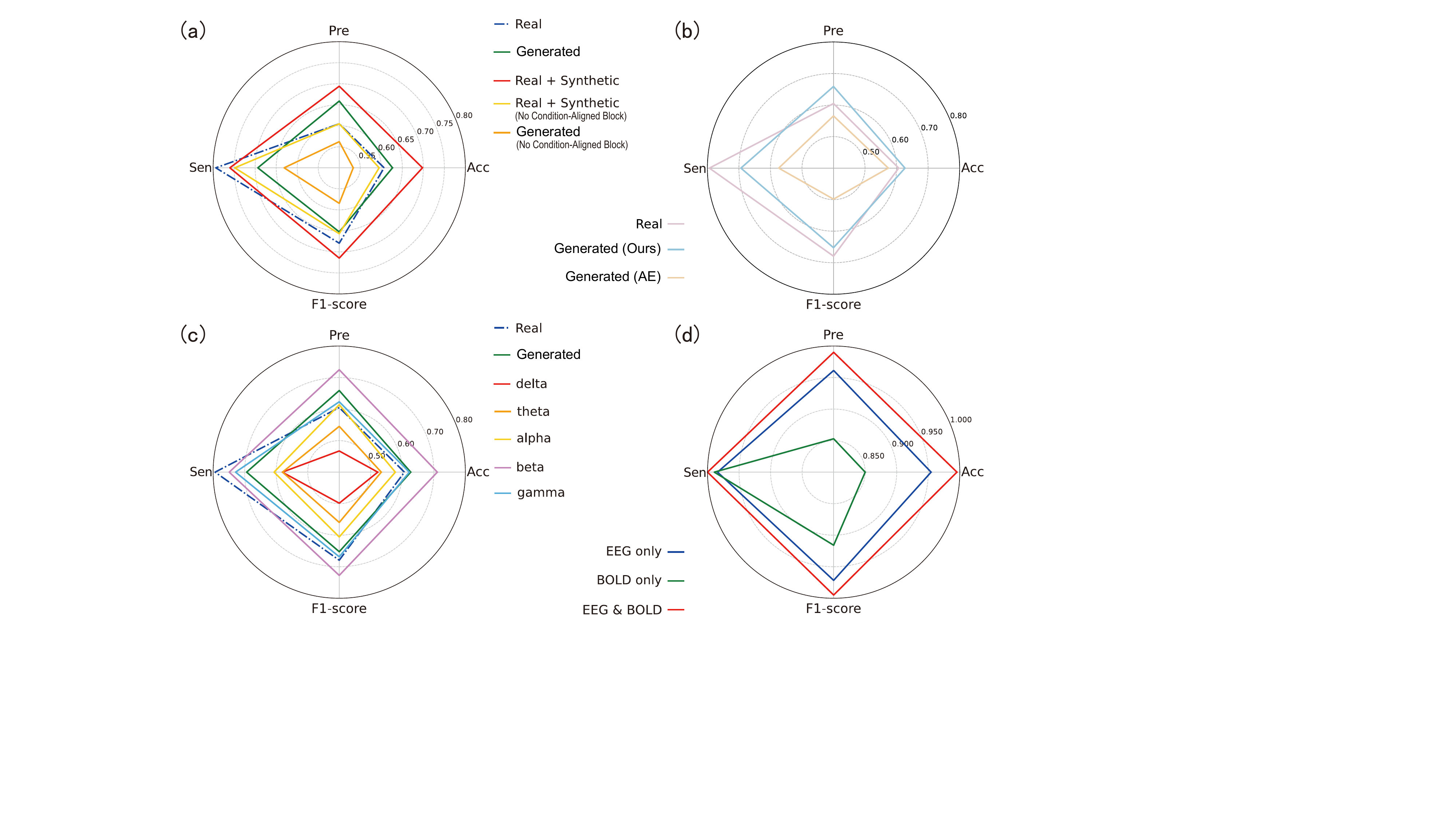}
	\caption{ (a) Radar plot comparing real BOLD signals, generated BOLD signals, their combination, and ablation results of conditioned blocks for motor imagery and resting state classification. (b) Radar plot of the prediction results of motor imagery and rest state for the real BOLD signal, the data generated by our method, and the data generated by the compared AE-based method. (c) Radar plot of BOLD signals synthesized from different EEG frequency bands in motor imagery and resting states using the proposed CATD framework. (d) Radar plot of BOLD signals synthesized from real EEG signals, and their combination, for predicting Parkinson’s disease in a clinical decision support experiment.}
	\label{fig3}
\end{figure*}

\subsubsection{Dataset}
The experiments were performed on the following datasets: the motor imagery dataset\cite{MI_lioi2020simultaneous}, the NODDI dataset\cite{noddi1_deligianni2016noddi, noddi2_deligianni2014relating}, and the EEG dataset of Parkinson's patients\cite{Parkinson_cavanagh2021eeg}. For evaluating the generated results, the XP1 part of the motor imagery dataset was used, which contained data from 10 subjects (2 females and 8 males, mean age: 28.4 years $\pm$ 10.6 years). Their paired 64-channel EEG and whole-brain fMRI scans were acquired simultaneously using a block design, with each block consisting of a 20-second rest and a 20-second motor imagery. The training set included data from 7 subjects (IDs xp101 to xp103, xp107 to xp110), while the test set consisted of data from 3 subjects (IDs xp104 to xp106). This division ensured the independence of the training and testing samples, providing a reliable assessment of the model's generalization performance across participants. The motor imagery task contained 48 blocks per subject during the test duration. The NODDI dataset included simultaneous resting-state 64-channel EEG and whole-brain fMRI scan data from 17 adult volunteers (11 males and 6 females, mean age: 32.84 $\pm$ 8.13 years). For evaluation, data from 2 participants (IDs 48 and 49) were selected as the test set, while the remaining data were used for training. This division allowed the model to accurately generate BOLD signals from EEG data. The EEG dataset of the Parkinson's patients contained 64-channel resting-state EEG data from 8 subjects (4 males and 4 females, mean age: 74.25 years $\pm$ 8.75 years). Due to the small sample size and significant clinical variability, data from 4 representative participants were selected based on factors such as age and gender for testing, highlighting the clinical potential of the method in real-world applications. EEG data in all datasets were obtained using the international 10-20 lead system.

\subsubsection{Implementation Detail}
The experiments were conducted on a server platform equipped with two NVIDIA Tesla A800 compute cards. For model parameters, the depth of the Transformer Blocks was set to 12, the hidden space dimension of the patch was 768, and the number of heads in the attention network was 12. Training was performed using the Adam optimizer with an initial learning rate of 0.0001, a batch size of 8, and 1000 epochs. To compute the experimental classification metrics, a five-fold cross-validation method was employed to ensure the reliability of the results. For calculating other quantitative metrics, five independent experiments were conducted, and the results were averaged to ensure data accuracy and stability.

\subsubsection{Metrics}

Variety of categorical metrics were employed, including Accuracy (ACC), Precision (PRE), Sensitivity (SEN), and F1-score, to demonstrate the performance of the synthesized results in downstream tasks. Additionally, to evaluate generation quality in the spatial dimension, Root Mean Square Error (RMSE) and Structural Similarity Index (SSIM) were used. For the temporal dimension, Cosine Similarity and Concordance Correlation Coefficient (CCC) were used to assess generation quality. For the temporal super-resolution experiments, Signal-to-Noise Ratio (SNR) was also used to evaluate the effectiveness of the synthesized signal, with SNR values being base-10 logarithms. To present the results more intuitively, graphical representations of classification and generation performance in different experiments were provided. For example, in the medical decision support experiment, the potential application of the method in medical tasks was illustrated through difference maps of BOLD function maps.

\subsection{Evaluation of the generated BOLD signal}

To assess the effectiveness of the EEG-to-BOLD signal generation, both synthetic and real BOLD signals were utilized to differentiate between subjects' performance in motor imagery and resting states. The brain activity of the subjects was divided into 20-second chunks for both resting and motor imagery tasks. A clear distinction between these states by the model indicated effective learning of brain activity patterns. As shown in Fig.\ref{fig3}(a), the cross-modal synthetic BOLD signals outperformed the real data in accuracy and precision. When combined with real data, the classification metrics improved significantly, although recall remained similar to the real data. Ablation experiments demonstrated that CAB enhances the model's ability to use EEG features to constrain BOLD signal synthesis, improving classification metrics compared to models without the CAB. At the same time, the results in Fig.\ref{fig3}(b) provide a direct comparison with the existing AE-based model, and the results show that the proposed CATD framework achieves superior performance across all classification metrics.

\begin{figure}
	\centering
	\includegraphics[width=\linewidth]{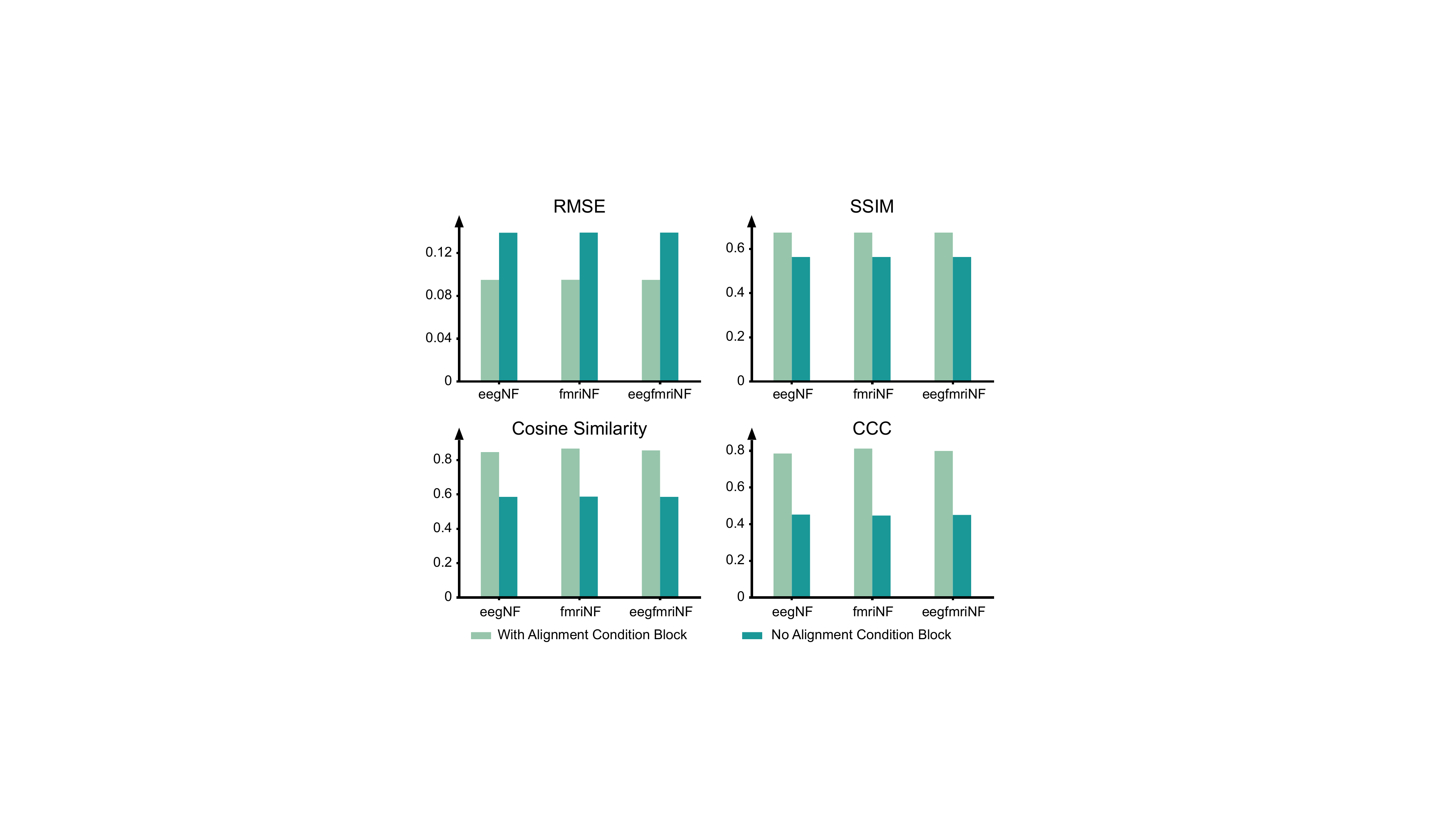}
	\caption{Results of quantitative spatial and temporal metrics for synthetic BOLD signals in three states of motor imagery from ablation experiments. The upper half shows spatial metrics, and the lower half shows temporal metrics, with light green indicating use of CAB and teal indicating no CAB.}
	\label{fig4}
\end{figure}

Spatial (RMSE, SSIM) and temporal (cosine similarity, CCC) metrics of the synthesized BOLD signal were also calculated for the three motor imagery states. The results in Fig.\ref{fig4} show low RMSEs (all below 0.1) and high SSIM (all above 0.6735). The cosine similarity is approximately 0.85, and the CCC value is about 0.8, indicating a high spatio-temporal correlation with the real signal. The ablation experiments (e.g., Fig.\ref{fig3}(a) and the No CAB section in Fig.\ref{fig4}) confirm the effectiveness of the CAB in improving signal quality.

\begin{figure*}
	\centering
	\includegraphics[width=0.85\linewidth]{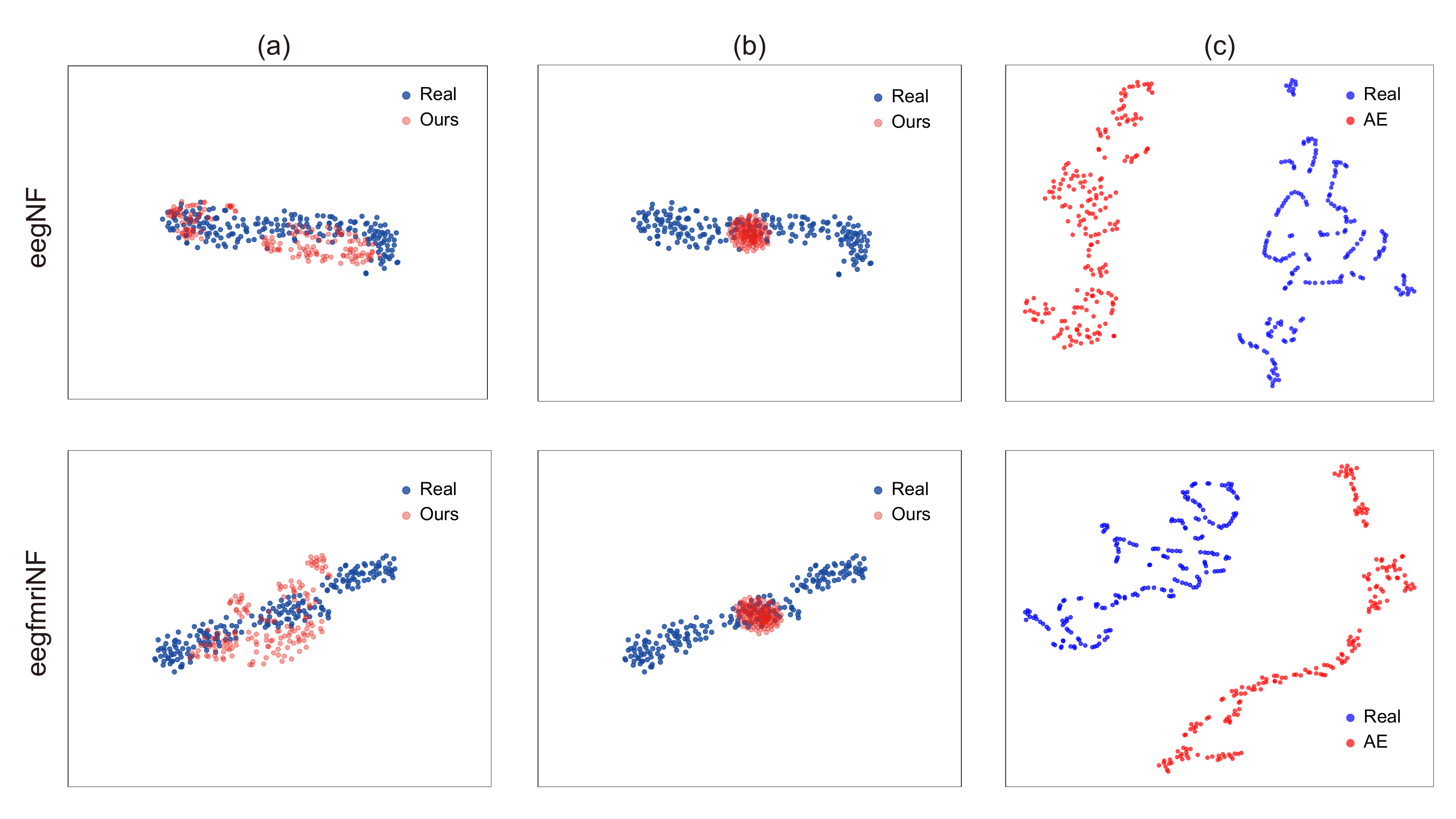}
	\caption{(a) t-SNE plots of synthesized versus real BOLD signal distributions in the CATD framework across two motor imagery states. (b) t-SNE plot of generated versus real BOLD signal distributions after ablation of the CAB. (c) t-SNE plots of BOLD signal distributions generated by the comparison method versus real signals.}
	\label{fig5}
\end{figure*}

To further illustrate the capability of the proposed model in learning high-dimensional heterogeneous brain activity features and achieving cross-modal transitions, t-SNE plots were used (Fig.\ref{fig5}). The distribution of BOLD signals synthesized with the full CATD framework more closely matches the real signal distribution, underscoring the critical role of the CAB in enhancing signal quality. As shown in Fig.\ref{fig5}(c), the comparison with the AE-based model demonstrates that the CATD framework not only generates a more accurate distribution of BOLD signals but also effectively captures the underlying structure of the real data, whereas the baseline method exhibits significant deviations from the real signal distribution. This further validates the advantage of the proposed approach in preserving the complex spatio-temporal relationships inherent in brain activity signals.

\subsection{Evaluation of temporal resolution enhanced BOLD signals}

To verify that the CATD framework can leverage the high temporal resolution of EEG signals to achieve the temporal super-resolution of BOLD signals, temporal resolution enhancement experiments were conducted. The proposed DTFS was used for EEG to achieve triple temporal super-resolution, meaning that the temporal resolution of the generated BOLD signal was three times that of the actual BOLD signal. Since the application scenario for temporal super-resolution typically involves enhancing the existing BOLD signal rather than generating it in the absence of a BOLD signal, the original low temporal resolution BOLD signal was used as a known condition. The enhanced high temporal resolution BOLD signal was obtained by constraining the generated signal using the low-resolution original BOLD signal. The high temporal resolution results were similarly evaluated in three different motor imagery states, as shown in Table \ref{tab1}.

\begin{table*}
	\centering
	\caption{Comparison of Time Series Similarity and SNR Between Generated and Real BOLD Signals Across Three Motor Imagery States}
	\label{tab1}
	\renewcommand\arraystretch{1.35}
	\resizebox{0.85\textwidth}{!}{%
		\begin{tabular}{cllll}
			\hline
			\multicolumn{1}{l}{} & \multicolumn{2}{c}{\textbf{Time Series Similarity}} & \multicolumn{2}{c}{\textbf{SNR}} \\ \hline
			\multicolumn{1}{l}{\textbf{}} &
			\multicolumn{1}{c}{\textbf{Cosine Similarity}} &
			\multicolumn{1}{c}{\textbf{CCC}} &
			\multicolumn{1}{c}{\textbf{Real}} &
			\multicolumn{1}{c}{\textbf{Synthetic}} \\ \hline
			\textbf{eegNF}       & 0.98998±0.00008          & 0.98589±0.00003          & 12.9640     & 12.9995±0.0103     \\
			\textbf{fmriNF}      & 0.99187±0.00007          & 0.98714±0.00006          & 13.4383     & 13.3112±0.0137     \\
			\textbf{eegfmriNF}   & 0.98740±0.00005          & 0.98209±0.00002          & 12.4715     & 13.0708±0.0533     \\ \hline
		\end{tabular}%
	}
\end{table*}

An important advantage of high temporal resolution is the potential for signal-to-noise ratio (SNR) enhancement. A higher SNR implies a greater proportion of useful information relative to noise, enabling signals with higher SNR to more reliably reflect actual physiological activity. In the experiments, the SNR values of real and generated BOLD signals were shown in Table \ref{tab1}. From the results, it could be seen that the signal-to-noise ratios in the two states were improved except for fmriNF, the motor imagery state. This was also consistent with the results demonstrated by t-SNE above, i.e., the results were worse in the fmriNF state and better in the other two states.

\begin{figure}
    \centering
    \includegraphics[width=\linewidth]{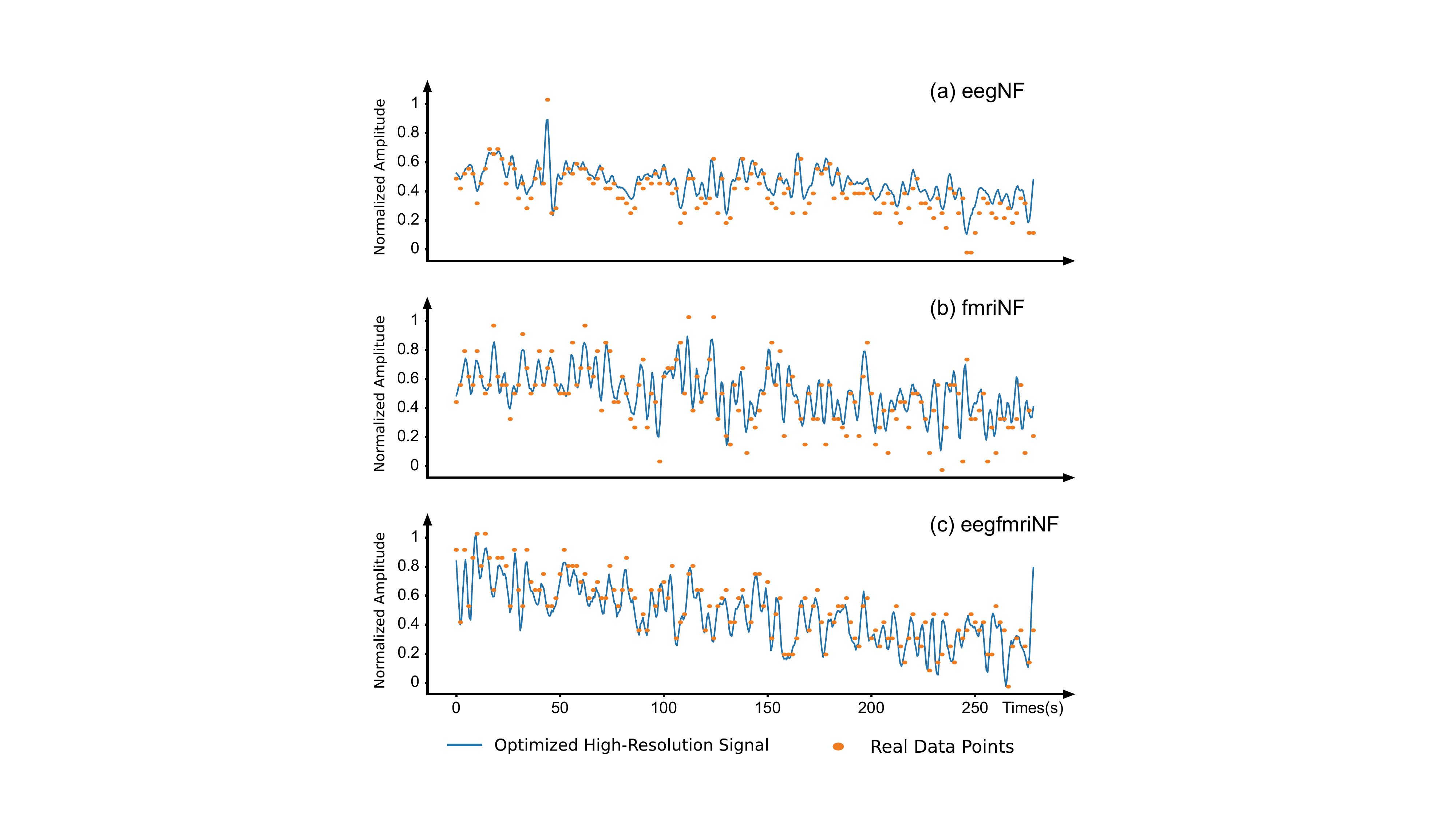}
    \caption{Visualization of low-resolution real BOLD signal points and high-resolution synthetic BOLD signal curves, demonstrating consistent trends.}
    \label{fig6}
\end{figure}

Data visualization was also performed. In the parietal region, where the correlation of motor imagery was strong, a node at the cortex corresponding to the C3 electrode portion of the international 10-20 lead system was selected. Curve graphs were used to display the signal intensities of the generated high temporal-resolution time-series signals and the real signals at this node at the corresponding time points. As shown in Fig.\ref{fig6}, the generated high temporal resolution signal and the real signal exhibited a high degree of similarity in trend. This further suggests that the high temporal resolution results obtained using the CATD framework can effectively reflect the trend of blood oxygenation.

\subsection{Evaluation of the effect of temporal features on the generated signals}

\begin{figure}
	\centering
	\includegraphics[width=\linewidth]{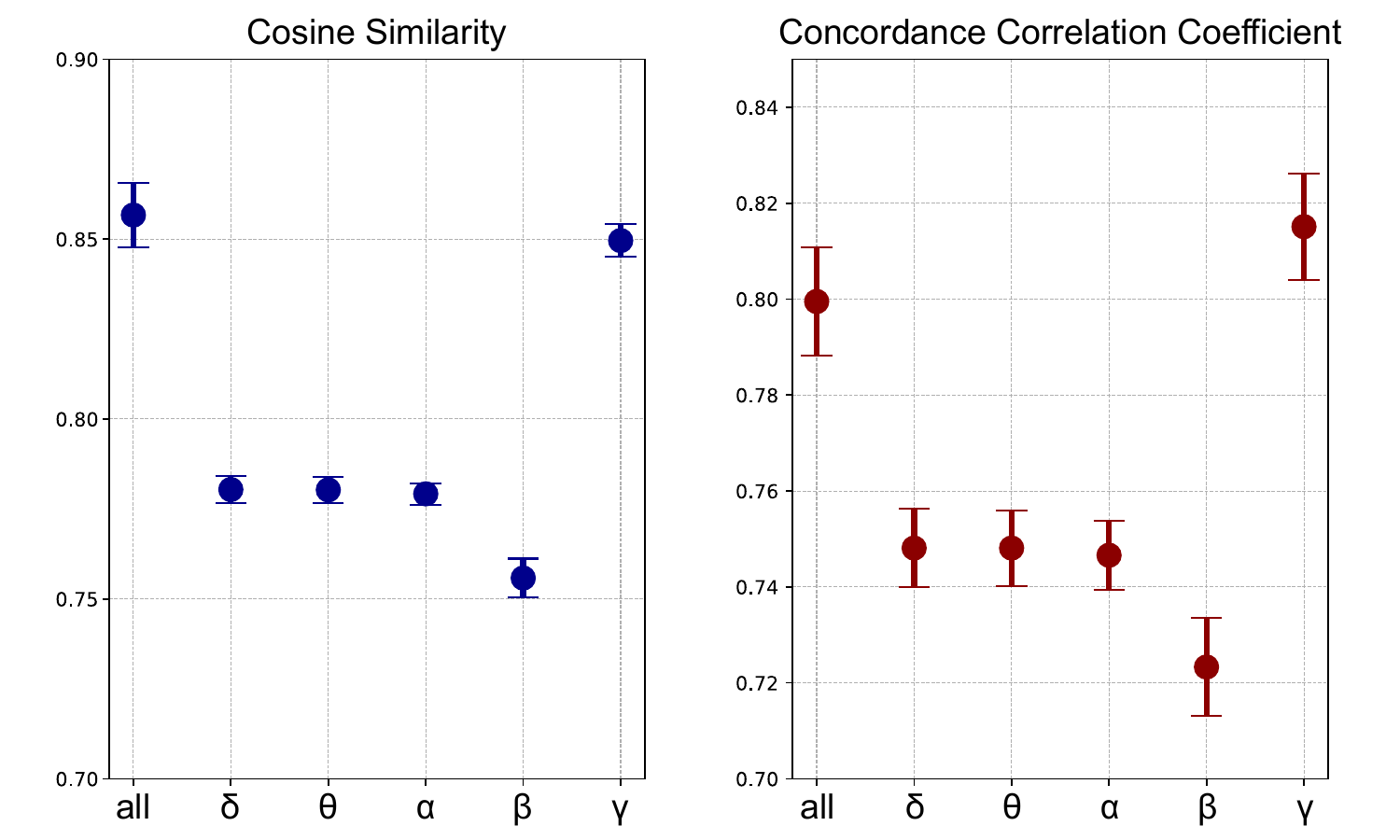}
	\caption{Cosine similarity and CCC comparison of the timing of synthesized BOLD signals with real BOLD signals using EEG signals from different frequency bands.}
	\label{fig7}
\end{figure}

Different frequency bands of EEG data reflect various types of brain activity, and these differences can significantly impact the results. The effect of EEG frequency bands on BOLD signal generated results was investigated. The EEG signals of different frequency bands were obtained by band-pass filtering, and BOLD signal generation was performed using these signals. As shown in Fig.\ref{fig3}(b), the beta and gamma bands, which are related to motor imagery, performed better in the classification metrics. Additionally, the temporal metrics of the generated results were calculated. As shown in Fig.\ref{fig7}, the gamma band performed well in temporal metrics, while the beta band exhibited lower temporal performance.

The reasons for these results are analysed below. The brain's state changes in preparation for or during imagined movements are generally slower and smoother, leading to event-related desynchronization (ERD) in the beta band\cite{ERD_pfurtscheller1999event}. This phenomenon may explain why the beta band excels in motor imagery and resting state categorization, even surpassing the full band's classification performance. However, this characteristic also causes the generated results of the beta band to differ considerably in temporal similarity, resulting in a lower temporal index compared to the blood oxygenation activity reflected by the full-band brain electrical activity, i.e., the real BOLD signal. On the other hand, the gamma band is closely associated with higher cognitive and perceptual functions. In motor imagery tasks, the brain's requirement to understand instructions and make judgments involves higher cognitive and perceptual functions \cite{Gamma_velasco2023motor}. Therefore, the gamma band most closely matches the real results in both the classification index and timing index. The other three low-frequency bands: delta, theta, and alpha, performed poorly on both the classification metric (e.g., Fig.\ref{fig3}(b)) and the temporal similarity metric (e.g., Fig.\ref{fig7}), suggesting that these bands negatively affect the generated results by containing less information about the correlation between the EEG and the BOLD signal.

\subsection{Support for medical decision-making}

\begin{figure*}
    \centering
    \includegraphics[width=\linewidth]{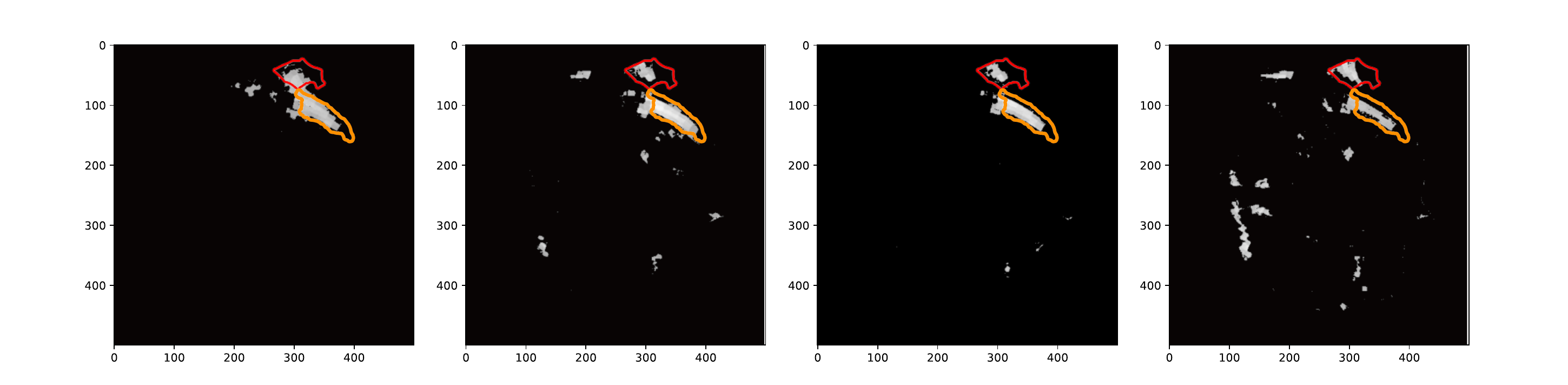}
    \caption{Comparison of difference maps of BOLD function maps synthesized using EEG from two healthy subjects and Parkinson's patients, with the portion marked by the red circle being the area of the brain that embodies the abnormality in both difference maps.}
    \label{fig8}
\end{figure*}

To demonstrate the potential application of the proposed CATD framework in the medical domain, cross-modal generation experiments were conducted on Parkinson's patients to support medical decision-making. In this part of the experiment, the framework was first trained on the NODDI dataset containing paired EEG and BOLD fMRI data from healthy subjects at resting state. Subsequently, a dataset containing only resting-state EEG data from Parkinson's patients was used to reconstruct their cortical BOLD functional maps using the trained cross-modal generation model.

Predictions of Parkinson's disease were performed using the recorded EEG signal, the generated BOLD signal, and their combination. The classification metrics are shown in Fig.\ref{fig3}(c). Results indicate that the simultaneous use of generated BOLD signals and recorded EEG signals significantly improves the accuracy of Parkinson's disease prediction and related metrics. This suggests that the CATD framework effectively captures the potential connection between BOLD signals and EEG signals and can be applied across different datasets, which is crucial for disease diagnosis. Consequently, the framework not only enhances the accuracy of existing diagnostic methods but also provides new tools and methodologies for the diagnosis and research of other neurological diseases.

In order to fully utilize the advantage of our CATD framework, i.e., to obtain BOLD signals with high spatial resolution without the condition of fMRI detection, we performed a difference analysis of generated BOLD functional maps in two healthy subjects and two Parkinson's patients. As shown in Fig.\ref{fig8}, both showed significant abnormalities in the brain regions marked by red circles. This region is the lingual gyrus, the medial occipito-temporal gyrus and the lateral occipito-temporal gyrus\cite{aparca2009s_destrieux2010automatic}, which, in patients with Parkinson's disease, shows significant structural changes\cite{pk_ev1_vignando2022mapping}, and in patients with PD accompanied by visual hallucinations, the atrophy of these three brain regions correlates with the severity of visual hallucinations\cite{pk_ev2_pagonabarraga2024parkinson}. This result demonstrates the potential application of the proposed method in localizing regions of abnormal brain activity and further proves its practical value in medical diagnosis.

\section{Discussion}\label{section4}

Cross-modal neuroimage synthesis is becoming crucial in neuroanalysis research. For the first time, we propose the CATD framework based on diffusion models to achieve cross-modal synthesis of temporal functional neuroimages, enabling the conversion of EEG to BOLD signals. This approach addresses limitations in BOLD acquisition, such as the inability of patients with metal implants to undergo fMRI scans. Enhancing BOLD temporal resolution facilitates the study of sub-millimeter cortical structures and activities\cite{18_raimondo2021line}, improves the quality and reliability of functional connectivity and task-driven fMRI research through higher sampling rates\cite{19_narsude2016three}, and allows for more accurate detection and localization of functional activation regions, capturing transient neural activities\cite{20_petrov2017improving}. This work provides novel insights for functional neuroimage synthesis\cite{21_yen2023exploring}. In the following, we will analyze the experimental results in detail to illustrate the performance of our framework.

In the experiments on the motor imagery dataset, our CATD framework demonstrates its superior performance in cross-modal feature learning and synthesis by detecting categorical metrics for both motor imagery states and resting states. Specifically, the framework is able to capture the features of EEG and BOLD signals well and perform effective cross-modal synthesis. This improved performance can be attributed to the ability to capture complex modes and integrate different modal features in the CATD framework, which enables it to achieve accurate transitions between various signal features. Ablation experiments further validate the effectiveness of the proposed CAB in aligning high-dimensional mismatched data pairs. CAB can learn and capture valid links between EEG and BOLD signals, ensuring the robustness of the model in cross-modal feature capture and synthesis. Compared to the AE-based model, the CATD framework achieves superior results due to the ability to extract advanced representations of brain activity and ensure better alignment of temporal and spatial features.

In the temporal super-resolution experiments, the results exhibit a high correlation in all three motor imagery states, while the signal-to-noise ratio is improved in two states compared to the original signal. This suggests that temporal resolution enhancement is indeed feasible and can provide a more detailed and accurate temporal representation of the BOLD signal.

In order to further verify which EEG frequency bands have a greater impact on the results and have the potential to improve the efficiency of cross-modal synthesis, a frequency band analysis was performed. It was found that the beta band had the greatest impact on the classification results, while the gamma band performed the best in terms of similarity to the real signal. This can be explained by the properties of EEG signals. Lower frequency bands such as beta show the event-related desynchronization (ERD) phenomenon in motor imagery tasks and therefore perform better in classification metrics. Whereas the higher cognitive functions involved in the motor imagery task are mainly determined by the gamma band, the generated results of the gamma band therefore show the highest similarity to the original signal.

The potential of the proposed method to enhance the accuracy of disease diagnosis was validated through disease decision support experiments. The results demonstrate that the method is capable of identifying potentially abnormal regions in the brain by leveraging the high spatial resolution of synthetic BOLD signals, even when only EEG signals are available. This capability contributes to improved diagnostic accuracy.

The CATD framework’s ability to synthesize BOLD signals from EEG has significant implications for clinical practice. By providing a non-invasive and cost-effective alternative to conventional fMRI, the framework can increase accessibility for patients who cannot undergo fMRI scans, such as those with metal implants. At the same time, the improved temporal resolution and functional maps generated by the framework can facilitate early diagnosis and intervention in neurological disorders like epilepsy and Parkinson’s disease. For example, identifying abnormal brain activity regions with high temporal and spatial resolution could help refine therapeutic strategies. The ability to monitor disease progression using synthetic neuroimages also offers potential for personalized medicine approaches.

An exciting direction for future research is the potential to extend the CATD framework to predict neural activity. Leveraging EEG signals and integrating temporal forecasting techniques within the CATD framework could enable the anticipation of upcoming BOLD signal patterns. While the current implementation does not achieve real-time generation due to the computational complexity of diffusion-based models, advancements in GPU and hardware acceleration technologies could enable real-time prediction in the near future. Real-time capabilities would significantly enhance the framework’s applicability in dynamic brain monitoring, real-time neurofeedback systems, and adaptive brain-machine interfaces.

\begin{figure}
    \centering
    \includegraphics[width=0.7\linewidth]{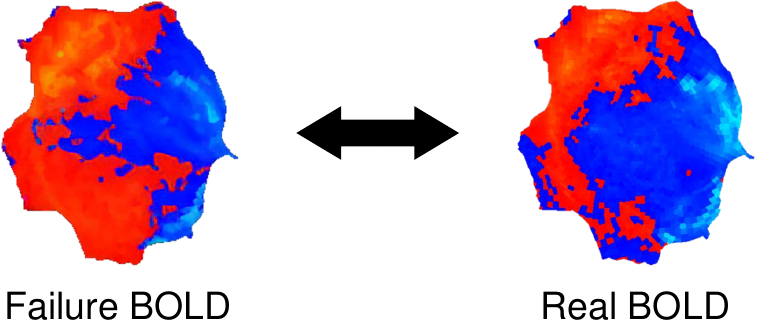}
    \caption{Typical failure case of our method, showing significant differences in the lower BOLD functional map (near the frontal lobe).}
    \label{fig9}
\end{figure}

While the results are promising, achieving accurate mapping from EEG to BOLD signals still presents challenges. Fig.\ref{fig9} illustrates a typical failure case, where the lower regions of the generated BOLD map significantly differ from the real image. This discrepancy is likely due to the low spatial resolution and relatively low signal-to-noise ratio (SNR) of EEG signals. Even though our model successfully learns the relationship between EEG and BOLD signals, the process of upsampling to generate high-resolution images introduces inaccuracies. Addressing these issues of spatial resolution and SNR will be crucial for enhancing the precision of cross-modal synthesis in future work.

In addition to the failure case highlighted in Fig.\ref{fig9}, other limitations exist. First, the limited number of paired data subjects constrained the training process. To address the limitations of dataset size, we have carefully selected, to the best of our knowledge, the relatively larger datasets among the publicly available ones with simultaneously acquired EEG and fMRI signals. However, the limited number of paired data subjects remains a constraint and poses challenges to achieving even greater robustness and generalizability. Introducing pre-trained models into the CATD framework could address the data scarcity issue by leveraging transfer learning to improve representation accuracy and enhance the model’s feature extraction capability. Second, inter-subject variability in EEG and BOLD signal patterns poses a challenge for achieving consistent cross-modal synthesis across diverse populations. Finally, the computational complexity of the CATD framework, particularly with its reliance on the CAB module and diffusion model, may hinder its scalability for large-scale clinical applications. Addressing these challenges in future work will be essential to fully realize the framework’s potential in both research and clinical settings.

\section{Conclusion}\label{section5}

In this work, a novel CATD framework is proposed for the cross-modal conversion of functional neuroimages, specifically the synthesis of BOLD signals from EEG signals. To fully exploit the high temporal resolution of EEG signals, the DTFS module was designed to increase the sampling rate of the EEG signal as a condition signal, achieving temporal resolution enhancement of the synthesized BOLD signal. By constructing the CAB module, the alignment of high-dimensional heterogeneous functional neuroimages in the hidden space was realized. Qualitative and quantitative experimental results demonstrate that the proposed framework effectively achieves cross-modal synthesis from EEG to BOLD signals. The effectiveness of CAB was validated through ablation experiments, and the framework's value was illustrated in practical application scenarios through medical decision support experiments. Future studies will focus on further optimizing the model and improving the quality of the generated signals to achieve more comprehensive functional neuroimage synthesis.

\bibliographystyle{ieeetr}
\bibliography{reference}

\end{document}